\documentclass[11pt]{article}
\usepackage{coling2020}
\usepackage{times}
\usepackage{latexsym}
\usepackage{graphicx} 
\usepackage{tabularx}
\usepackage{url}
\usepackage{graphicx}
\usepackage{amsmath}
\usepackage{amssymb}
\usepackage{booktabs} 
\usepackage{enumitem}
\usepackage{xcolor}
\usepackage{subcaption}

\newcommand{\cev}[1]{\reflectbox{\ensuremath{\vec{\reflectbox{\ensuremath{#1}}}}}}

\definecolor{forestgreen(traditional)}{rgb}{0.0, 0.27, 0.13}
\definecolor{falured}{rgb}{0.5, 0.09, 0.09}

\colingfinalcopy 

\title{Reinforced Multi-task Approach for  Multi-hop Question Generation }

\author{Deepak Gupta, Hardik Chauhan\thanks{\enspace Work done during an internship at IIT Patna.}, Akella Ravi Tej\footnotemark[1], Asif Ekbal, and Pushpak Bhattacharyya \\
  Indian Institute of Technology Patna, India  \\
  {\tt 
  \{deepak.pcs16,asif,pb\}@iitp.ac.in}\\
  {\tt \{chauhanhardik23,ravitej.akella\}@gmail.com
  }
  }

\date{}

\begin{document}
\maketitle
\begin{abstract}
Question generation (QG) attempts to solve the inverse of question answering (QA) problem by generating a natural language question given a \textit{document} and an \textit{answer}. While sequence to sequence neural models surpass rule-based systems for QG, they are limited in their capacity to focus on more than one supporting fact. For QG, we often require multiple supporting facts to generate high-quality questions. Inspired by recent works on multi-hop reasoning in QA, we take up Multi-hop question generation, which aims at generating relevant questions based on supporting facts in the context. We employ multitask learning with the auxiliary task of \textit{answer-aware} supporting fact prediction to guide the question generator. In addition, we also proposed a \textit{question-aware} reward function in a Reinforcement Learning (RL) framework to maximize the utilization of the supporting facts. We demonstrate the effectiveness of our approach through experiments on the multi-hop question answering dataset, HotPotQA. Empirical evaluation shows our model to outperform the single-hop neural question generation models on both automatic evaluation metrics such as BLEU, METEOR, and ROUGE, and human evaluation metrics for quality and coverage of the generated questions.
\end{abstract}
\section{Introduction}
In natural language processing (NLP), question generation is considered to be an important yet challenging problem. Given a passage and answer as inputs to the model, the task is to generate a semantically coherent question for the given answer. 

In the past, question generation has been tackled using rule-based approaches such as question templates \cite{lindberg2013generating} or utilizing named entity information and predictive argument structures of sentences \cite{chali2015towards}. 
Recently, neural-based approaches have accomplished impressive results \cite{learning-to-ask,ans-focused-qg,kim2018improving,gupta2019improving} for the task of question generation.  The availability of large-scale machine reading comprehension datasets such as SQuAD \cite{squad}, NewsQA \cite{trischler2016newsqa}, MSMARCO \cite{ms-marco} etc. have facilitated 
research in question answering task. These large-scale datasets have also been used to create the resources \cite{singh2019xlda,GUPTA18.826} and evaluate the performance \cite{asai2018multilingual,gupta2018uncovering,gupta2019deep} of the system in low-resource reading comprehension task.  SQuAD dataset itself has been the de facto choice for most of the previous works in question generation. However, 90\% of the questions in SQuAD can be answered from a single sentence \cite{min-etal-2018-efficient}, hence former QG systems trained on SQuAD are not capable of distilling and utilizing information from multiple sentences. Recently released multi-hop datasets such as \texttt{QAngaroo} \cite{qangaroo}, \texttt{ComplexWebQuestions} \cite{complex-web-qa-dataset} and \texttt{HotPotQA} \cite{hotpot} are more suitable for building QG systems that required to gather and utilize information across multiple documents as opposed to a single paragraph or sentence.\\
\begin{table}[h]
\centering
\begin{tabularx}{\linewidth}{X}
\toprule
\textbf{Document:} \textcolor{blue}{ A few sects, such as the} \textcolor{red}{Bishnoi}, \textcolor{blue}{lay special emphasis on the conservation of particular species, such as the antelope. } \textbf{\textit{(ii)}} ... \\
\textbf{Question$_{SHQ}$:} \textit{Who lay special emphasis on conservation of particular species ?}\\
\hline
\textbf{Document (1):} 
\textbf{\textit{(i)}} \textcolor{blue}{Stig Lennart Blomqvist (born 29 July 1946) is a Swedish rally driver. }\\
\textbf{\textit{(ii)}}  ...
\textbf{\textit{(iii)}} \textcolor{blue}{Driving an Audi Quattro for the Audi factory team, Blomqvist won the World Rally Championship drivers' title in 1984 and finished runner-up in 1985. }\\
\textbf{Document (2):} \textbf{\textit{(i)}} \textcolor{red}{ The Audi Quattro} \textcolor{blue}{ is a road and rally car, produced by the German automobile manufacturer Audi, part of the Volkswagen Group. } \textbf{\textit{(ii)}} ... \\
\textbf{Question$_{MHQ}$:} \textit{Which car produced by German automobile manufacturer, was driven by Stig Lennart Blomqvist?}\\
\hline
\end{tabularx}
\caption{An example of Single-hop question (SHQ) from the \texttt{SQuAD} dataset and a Multi-hop Question (MHQ) from the \texttt{HotPotQA} dataset. The relevant sentences and answer required to form the question are highlighted in \textcolor{blue}{blue} and \textcolor{red}{red} respectively.}
\label{table:intro_example}
\end{table}
\indent In multi-hop question answering, one has to reason over multiple relevant sentences from different paragraphs to answer a given question. We refer to these relevant sentences as supporting facts in the context. Hence, we frame \textit{Multi-hop question generation} as the task of generating the question conditioned on the information gathered from reasoning over all the supporting facts across multiple paragraphs/documents.  Since this task requires assembling and summarizing information from multiple relevant documents in contrast to a single sentence/paragraph, therefore, it is more challenging than the existing single-hop QG task. Further, the presence of irrelevant information makes it difficult to capture the supporting facts required for question generation. The explicit information about the supporting facts in the document is not often readily available, which makes the task more complex. In this work, we provide an alternative to get the supporting facts information from the document with the help of multi-task learning. Table \ref{table:intro_example} gives sample examples from SQuAD and HotPotQA dataset. It is cleared from the example that the single-hop question is formed by focusing on a single sentence/document and answer,  while in  multi-hop question, multiple supporting facts from different documents and answer are accumulated to form the question.\\
\indent Multi-hop QG has real-world applications in several domains, such as education, chatbots, etc. The questions generated from the multi-hop approach will inspire critical thinking in students by encouraging them to reason over the relationship between multiple sentences to answer correctly. Specifically, solving these questions requires higher-order cognitive-skills (e.g., applying, analyzing). Therefore, forming challenging questions is crucial for evaluating a student’s knowledge and stimulating self-learning. Similarly, in goal-oriented chatbots, multi-hop QG is an important skill for chatbots, e.g., in initiating conversations, asking and providing detailed information to the user by considering multiple sources of information. In contrast, 
in a single-hop QG, 
only single source of information is considered while generation.

In this paper, we propose to tackle Multi-hop QG problem in two stages. 
In the first stage, we learn supporting facts aware encoder representation to predict the supporting facts from the documents by jointly training with question generation and  subsequently enforcing the utilization of these supporting facts. The former is achieved by sharing the encoder weights with an \textit{answer-aware} supporting facts prediction network, trained jointly in a multi-task learning framework. The latter objective is formulated as a \textit{question-aware} supporting facts prediction reward, which is optimized alongside supervised sequence loss. Additionally, we observe that multi-task framework offers substantial improvements in the performance of question generation and also avoid the inclusion of noisy sentences information in generated question, and reinforcement learning (RL) brings the complete and complex question  to otherwise maximum likelihood estimation (MLE) optimized QG model. 
Our main contributions in this work are: \textbf{(i).} We introduce the problem of multi-hop question generation and propose a multi-task training framework to condition the shared encoder with supporting facts information.
\textbf{(ii).}
We formulate a novel reward function, multihop-enhanced reward via \textit{question-aware} supporting fact predictions to enforce the maximum utilization of supporting facts to generate a question; 
\textbf{(iii).} 
We introduce an automatic evaluation metric to measure the coverage of supporting facts in the generated question. \textbf{(iv).} Empirical results show that our proposed method outperforms the current state-of-the-art single-hop QG models over several automatic and human evaluation metrics on the HotPotQA dataset.

\section{Related Work}

Question generation literature can be broadly divided into two classes based on the features used for generating questions. The former regime consists of rule-based approaches \cite{heilman2010good,chali2015towards} that rely on human-designed features such as named-entity information, etc. to leverage the semantic information from a context for question generation. In the second category, question generation problem is treated as a sequence-to-sequence \cite{sutskever2014seq} learning problem, which involves automatic learning of useful features from the context by leveraging the sheer volume of training data. 
The first neural encoder-decoder model for question generation was proposed in \newcite{learning-to-ask}. However, this work does not take the answer information into consideration while generating the question.
Thereafter, several neural-based QG approaches \cite{ans-focused-qg,para-level-max-out-qg,chen2018learningq-qg} have been proposed that utilize the answer position information and copy mechanism. \newcite{DBLP:journals/corr/WangYT17} and \newcite{guo-etal-2018-soft} demonstrated an appreciable improvement in the performance of the QG task when trained in a multi-task learning framework.

The model proposed by \newcite{Seo2017Bidirectional} and \newcite{weissenborn-etal-2017-making} for single-document QA experience a significant drop in accuracy when applied in multiple documents settings. This shortcoming of single-document QA datasets is addressed by newly released multi-hop datasets \cite{qangaroo,complex-web-qa-dataset,hotpot} that promote multi-step inference across several documents. So far, multi-hop datasets have been predominantly used for answer generation tasks \cite{seo2016query,tay2018multi,zhang2018variational}. Our work can be seen as an extension to single hop question generation where a non-trivial number of supporting facts are spread across multiple documents.
\section{Proposed Approach}
\begin{figure*}
    \centering
    \includegraphics[width=\linewidth]{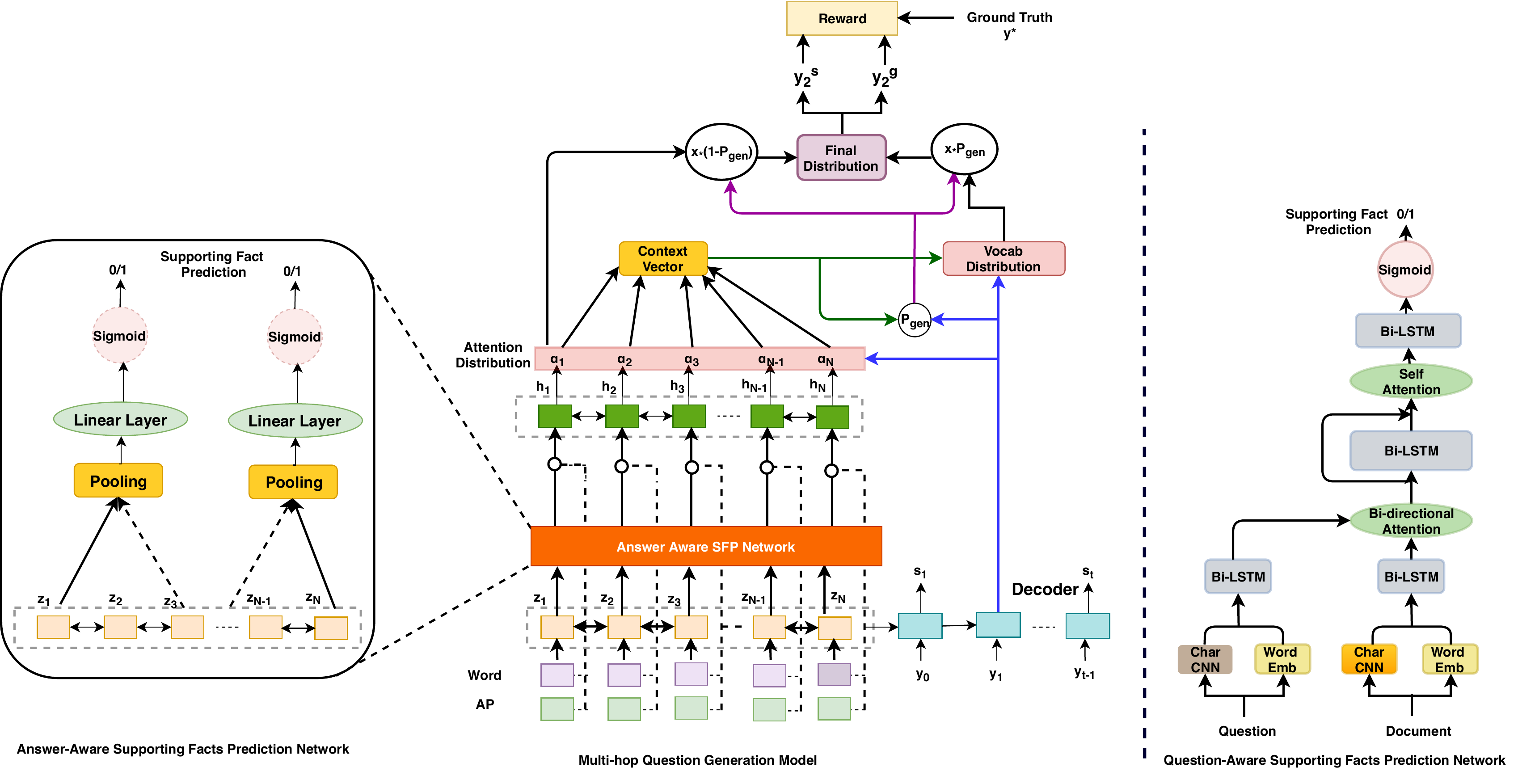}
    \caption{
    Architecture of our proposed Multi-hop QG network. The inputs to the model are the document word embeddings and answer position (\textbf{AP}) features. Question generation and answer-aware supporting facts prediction model (\textbf{left}) jointly train the shared document encoder (Bi-LSTM) layer. The image on the \textbf{right} depicts our question-aware supporting facts prediction network, which is our  MultiHop-Enhanced Reward function. The inputs to this model are the generated question (output of multi-hop QG network) and a list of documents. 
    }
    \label{fig:proposed-model}
\end{figure*}

\paragraph{Problem Statement:}
In multi-hop question generation, we consider a document list $L$ with $n_L$ documents, and an $m$-word answer $A$. Let the total number of words in all the documents $D_i \in L$ combined be $N$. 
Let a document list $L$ contains a total of $K$ candidate sentences $CS=\{S_1, S_2, \ldots, S_K\}$ and a set of supporting facts\footnote{It is only used to train the network. At the time of testing, network predict the supporting facts to be used for question generation.} $SF$ such that $SF \in CS$. The answer $A=\{w_{D_k^{a_1}} , w_{D_k^{a_2}}, \ldots, w_{D_k^{a_m}} \}$ is an $m$-length text span in one of the documents $D_k \in L$. Our task is to generate an $n_Q$-word question sequence $\hat{Q}= \{y_1, y_2, \ldots, y_{n_Q} \}$ whose answer is based on the supporting facts $SF$ in document list $L$. Our proposed model for multi-hop question generation is depicted in Figure \ref{fig:proposed-model}.


\subsection{Multi-Hop Question Generation  Model} \label{answer-focused} In this section, we discuss the various components of our proposed Multi-Hop QG model. Our proposed model has four components \textit{(i). Document and Answer Encoder} which encodes the list of documents and answer to further generate the question, \textit{(ii). Multi-task Learning} to facilitate the QG model to automatically select the supporting facts to generate the question, \textit{(iii). Question Decoder}, which generates questions using the pointer-generator mechanism and \textit{(iv). MultiHop-Enhanced QG} component which forces the model to generate those questions which can maximize the supporting facts prediction based reward. 

\subsubsection{Document and Answer Encoder} The encoder of the Multi-Hop QG model encodes the answer and documents using the layered Bi-LSTM network.
\paragraph{Answer Encoding:}
We introduce an answer tagging feature that encodes the relative position information of the answer in a list of documents. The answer tagging feature is an $N$ length list of vector of dimension $d_1$, where each element has either a tag value of $0$ or $1$. Elements that correspond to the words in the answer text span have a tag value of $1$, else the tag value is $0$. We map these tags to the embedding of dimension $d_1$. 
We represent the answer encoding features using $\{a_1, \ldots, a_N\}$. 
\paragraph{Hierarchical Document Encoding:}
\label{doc_enc}
To encode the document list $L$, we first concatenate all the documents $D_k \in L$, resulting in a list of $N$ words. Each word in this list is then mapped to a $d_2$ dimensional word embedding $u \in \mathbb{R}^{d_2}$. We then concatenate the document word embeddings with answer encoding features and feed it to a bi-directional LSTM encoder $\{LSTM^{fwd}, LSTM^{bwd}\}$.
\begin{equation}
    z_t= LSTM(z_{t-1}, [u_t, a_t])
\label{lstm-1st-layer}
\end{equation}
We compute the forward hidden states $\vec{z}_{t}$ and the backward hidden states $ \cev{z}_{t}$ and concatenate them to get the final hidden state $z_{t} = [\vec{z}_{t} \oplus \cev{z}_{t}]$.
The answer-aware supporting facts predictions network (will be introduced shortly) takes the encoded representation as input and predicts whether the candidate sentence is a supporting fact or not. We represent the predictions with $p_1, p_2, \ldots, p_K$. Similar to answer encoding, we map each prediction $p_i$ with a vector $v_i$ of dimension $d_3$. 

A candidate sentence $S_i$ contains the $n_i$ number of words. In a given document list $L$, we have $K$ candidate sentences such that $\sum_{i=1}^{i=K} n_i = N$. We generate the supporting fact encoding $sf_i \in \mathbb{R}^{n_i \times d_3}$ for the candidate sentence $S_i$ as follows:
\begin{equation}
    sf^i =  e_{n_i}v_i^\text{T}
\end{equation}
where $e_{n_i} \in \mathbb{R}^{n_i}$ is a vector of 1s. The rows of $sf_i$ denote the supporting fact encoding of the word present in the candidate sentence $S_i$. 
We denote the supporting facts encoding of a word $w_t$ in the document list $L$ with $s_t \in \mathbb{R}^{d_3}$.
Since, we 
also deal with the answer-aware supporting facts predictions in a multi-task setting, therefore, to obtain a supporting facts induced encoder representation, we introduce another Bi-LSTM layer.
\begin{equation}
    h_t= LSTM(h_{t-1}, [z_t, u_t, a_t, s_t])
    \label{lstm-2nd-layer}
\end{equation}
Similar to the first encoding layer, we concatenate the forward and backward hidden states to obtain the final hidden state representation. 

\subsubsection{Multi-task Learning for Supporting Facts Prediction}
We introduce the task of \textit{answer-aware supporting facts prediction} to condition the QG model's encoder with the supporting facts information.  Multi-task learning 
facilitates the QG model to automatically select the supporting facts conditioned on the given answer. This is achieved by using a multi-task learning framework where the \textit{answer-aware} supporting facts prediction network and Multi-hop QG share a common document encoder (Section \ref{doc_enc}). The network takes the encoded representation of each candidate sentence $S_i \in CS$ as input and sentence-wise predictions for the supporting facts. More specifically, we concatenate the first and last hidden state representation of each candidate sentence from the encoder outputs and pass it through a fully-connected layer that outputs a Sigmoid probability for the sentence to be a supporting fact. The architecture of this network is illustrated in Figure \ref{fig:proposed-model} (\textbf{left}). This network is then trained with a binary cross entropy loss and the ground-truth supporting facts labels:
\begin{equation}
\small
\mathcal{L}_{sp} = - \frac{1}{N}\sum_{j=1}^{N}\sum_{i=1}^{n_j} \delta_{y_i^j=1}\log (p_i^{j}) + (1-\delta_{y_i^j\neq 1})\log (1-p_i^{j})
\label{eq:sp-loss} 
\end{equation}
where $N$ is the number of document list, $S$ the number of candidate sentences in a particular training example, $\delta_i^j$ and $p_i^{j}$ represent the ground truth supporting facts label and the output Sigmoid probability, respectively.
\subsubsection{Question Decoder}
\label{decoder}
We use a LSTM network with global attention mechanism \cite{global-attention} to generate the question $\hat{Q} = \{y_1, y_2, \ldots, y_m\}$ one word at a time. We use copy mechanism \cite{go-to-the-point-see-2017,pointing-unknown-gulcehre-2016} to deal with rare or unknown words. At each timestep $t$,
\begin{equation}
    s_t= LSTM(s_{t-1}, y_{t-1})
\end{equation}
The attention distribution $\alpha_t$ and context vector $c_t$ are obtained using the following equations:
\begin{equation}
    \begin{aligned}
    e_{t,i}&= s_t^{T}*h_i\\
    \alpha_{t,i} &= \frac{\exp (e_{t,i})}{\sum_{j=1}^{N}\exp (e_{t,j})}\\
    c_{t} &= \sum_{i = 1}^{N} \alpha_{t,i}h_{i}
     \end{aligned}
\end{equation}
The probability distribution over the question vocabulary is then computed as,
\begin{equation}
    P_{vocab} = \text{softmax}(\text{tanh}(\mathbf{W_q}*[c_t\oplus s_t]))
\end{equation}
where $\mathbf{W_q}$ is a weight matrix. The probability of picking a word (generating) from the fixed vocabulary words, or the probability of not copying a word from the document list $L$ at a given timestep $t$ is computed by the following equation:
\begin{equation}
    P_{gen} = 1 - \sigma (\mathbf{W_a}c_t+ \mathbf{W_b}s_t)
\end{equation}
where, $\mathbf{W_a}$ and $\mathbf{W_b}$ are the weight matrices and $\sigma$ represents the Sigmoid function. The probability distribution over the words in the document is computed by 
summing over all the attention scores of the corresponding words:
\begin{equation}
    P_{copy}(w) = \sum_{i=1}^{N} \alpha_{t, i} * \mathbf{1}\{w==w_i\}
\end{equation}
where $\mathbf{1}\{w==w_i\}$ denotes the vector of length $N$ having the value $1$ where $w==w_i$, otherwise $0$. The final probability distribution over the dynamic vocabulary (document and question vocabulary) is calculated by the following:
\begin{equation}
\footnotesize
    P(w) = P_{gen} * P_{vocab}(w) + (1 - P_{gen}) * P_{copy}(w)
\end{equation}
 \subsubsection{MultiHop-Enhanced  QG}
 \label{mer-reward}
We introduce a reinforcement learning based reward function and sequence training algorithm to train the RL network. The proposed reward function forces the model to generate those questions which can maximize the reward. 
\paragraph{MultiHop-Enhanced Reward (MER):}
Our reward function is a neural network, we call it
\textit{Question-Aware Supporting Fact Prediction } network. 
We train our neural network based reward function for the supporting fact prediction task on \texttt{HotPotQA} dataset. This network takes as inputs the list of documents $L$ and the generated question $\hat{Q}$, and predicts the supporting fact probability for each candidate sentence. 
This model subsumes the latest technical advances of question answering, including character-level models, self-attention \cite{wang2017gated}, and bi-attention \cite{Seo2017Bidirectional}. The network architecture of the supporting facts prediction model is similar to \newcite{hotpot}, as shown in Figure \ref{fig:proposed-model} (\textbf{right}). 
For each candidate sentence in the document list, we concatenate the output of the self-attention layer at the first and last positions, and use a binary linear classifier to predict the probability that the current sentence is a supporting fact. This network is pre-trained on HotPotQA dataset using binary cross-entropy loss. \\
\indent For each generated question, we compute the F1 score (as a reward) between the ground truth supporting facts and the predicted supporting facts. This reward is supposed to be carefully used because the QG model can cheat by greedily copying words from the supporting facts to the generated question. In this case, even though high MER is achieved, the model loses the question generation ability. To handle this situation, we regularize this reward function with additional Rouge-L reward, which avoids the process of greedily copying words from the supporting facts by ensuring the content matching between the ground truth and generated question. We also experiment with BLEU as an additional reward, but Rouge-L as a reward has shown to outperform the BLEU reward function (Please see the \textbf{Appendix} for the results).
\paragraph{Adaptive Self-critical Sequence Training:}
We use the REINFORCE \cite{williams1992simple} algorithm to learn the policy defined by question generation model parameters, which can maximize our expected rewards. To avoid the high variance problem in the REINFORCE estimator, self-critical sequence training (SCST) \cite{rennie2017self} framework is used for sequence training that uses greedy decoding score as a baseline. 
In SCST, during training, two output sequences are produced: $y^{s}$, obtained by sampling from the probability distribution $P(y^s_t | y^s_1, \ldots, y^s_{t-1}, \mathcal{D})$, and $y^g$, the greedy-decoding output sequence. We define $r(y,y^*)$ as the reward obtained for an output sequence $y$, when the ground truth sequence is $y^*$. The SCST loss can be written as,
\begin{equation}
\small
\begin{split}
\mathcal{L}_{rl}^{scst} & = - (r(y^{s}, y^*) - r(y^{g},y^*)) * R
\end{split}
\label{eq:rl-loss}
\end{equation}
where, $R= \sum_{t=1}^{n'} \log P(y^s_t | y^s_1,  \ldots, y^s_{t-1}, \mathcal{D}) $.
However, the greedy decoding method only considers the single-word probability, while the sampling
considers the probabilities of all words in the vocabulary. Because of this the greedy reward $r(y^{g},y^*)$ has higher variance than the Monte-Carlo sampling reward $r(y^{s}, y^*)$, and their gap is also very unstable. We experiment with the SCST loss and observe that greedy strategy causes SCST to be unstable in the training progress. Towards this, we introduce a weight history factor similar to \cite{zhu2018image}. The history factor is the ratio of the mean sampling reward and mean greedy strategy reward in previous $k$ iterations. We update the SCST loss function in the following way:
\begin{equation}
\small
\begin{split}
\mathcal{L}_{rl} & = - \big(r(y^{s}, y^*) - \alpha \frac{\sum_{i=t-h+1}^{i=t} r_i(y^{s}, y^*)}{\sum_{i=t-h+1}^{i=t} r_i(y^{g}, y^*)} r(y^{g},y^*) \big)  
\times R
\end{split}
\label{eq:ascst-loss}
\end{equation}
where $\alpha$ is a hyper-parameter,  $t$ is the current iteration, $h$ is the history determines, the number of previous rewards are used to estimate. The denominator of the history factor is used to normalize the current greedy reward $ r(y^{g},y^*)$ with the mean greedy reward of previous $h$ iterations. The numerator of the history factor ensures the greedy reward has a similar magnitude with the mean sample reward of previous $h$ iterations.
\\

\begin{table*}[h] 
\begin{center}
\resizebox{0.9\textwidth}{!}{%
\begin{tabular}{l|l|l|l|l|l|l|l}
\hline
\textbf{Model} & \textbf{BLEU-1} & \textbf{BLEU-2} & \textbf{BLEU-3} & \textbf{BLEU-4} & \textbf{METEOR} & \textbf{ROUGE-L} & \textbf{SF Coverage} \\ \hline
\textit{s2s} \cite{learning-to-ask} & 34.98 & 22.55 & 16.79 & 13.25 & 17.58 &  33.75& 59.61  \\ 
\textit{s2s+copy} & 31.86 & 22.47 & 17.70 & 14.63 & 19.47 & 30.45 & 60.48 \\ 
\textit{s2s+answer} & 39.63 & 27.35 & 21.00 & 16.83 & 17.58 & 33.75 & 61.26 \\
\textit{NQG} \cite{zhou2017neural-qg} & 39.82 & 29.24 & 23.45 & 19.55 & 21.39 & 36.63 & 61.55\\ 
\textit{ASs2s} \cite{kim2018improving} & 39.08 & 29.06 & 23.45 & 19.66 & 22.84 & 36.98 &  64.22\\ 
\textit{Max-out Pointer} \cite{para-level-max-out-qg} & 42.58 & 30.91 & 24.61 & 20.39 & 20.36 & 35.31 & 63.93 \\ 

\textit{Semantic-Reinforced} \cite{zhang-bansal-2019-addressing} & 44.07 & 32.72 & 26.18 & 21.69 & \textbf{23.61} & 39.40 &68.74 \\ \hline
\textit{SharedEncoder-QG} (Ours)& 41.72 & 30.75 & 24.72 & 20.64 & 22.01 &37.18 & 65.46 \\
\textit{MTL-QG} (Ours) & 44.17 & 32.34 & 25.74 & 21.28 & 21.21 &37.55 &  70.11\\ \hline
Proposed Model & \textbf{46.80} & \textbf{34.94} &\textbf{ 28.21} & \textbf{23.57 }& {22.88} & \textbf{39.68} & \textbf{ 74.37}\\ \hline
\hline

\end{tabular}%
}
\end{center}
\caption{Performance comparison between proposed approach and state-of-the-art QG models on the test set of \texttt{HotPotQA}. Here \textbf{\textit{s2s}}: sequence-to-sequence, \textbf{\textit{s2s+copy}}: s2s with copy mechanism \cite{go-to-the-point-see-2017}, \textbf{\textit{s2s+answer}}: s2s with answer encoding.}
\label{main-results}
\end{table*}

\section{Experimental Setup}\label{exp-setup}
\subsection{Network Training}
We train our multi-task learning based question generator model (\textit{MTL-QG}) using ``teacher forcing'' algorithm \cite{williams1989learning} to minimize the negative log-likelihood of the model on the training data. With $y^* = \{y^*_1, y^*_2, \ldots, y^*_{m}\}$ as the ground-truth output sequence for a given input sequence $D$, the maximum-likelihood training objective can be written as,
\begin{equation}
\mathcal{L}_{ml} = -\sum_{t=1}^{m} \log P(y^*_t | y^*_1, \ldots, y^*_{t-1}, \mathcal{D})
\label{eq:ml-loss}
\end{equation}
Since, we are jointly training the supporting-fact prediction network with question generation network then the total loss for the \textit{MTL-QG} network is $\mathcal{L}_{ml} + \beta \mathcal{L}_{sp}$. We use a mixed-objective learning function \cite{wu2016google,paulus2017deep} to train the final network:
\begin{equation}
\small
\mathcal{L}_{mixed} = \gamma_1 \mathcal{L}_{rl} + \gamma_2 \mathcal{L}_{ml} +  \gamma_3 \mathcal{L}_{sp},
\label{eq:mixed-loss}
\end{equation}
where $\gamma_1$, $\gamma_2$, and $\gamma_3$ correspond to the weights of $\mathcal{L}_{rl}$, $\mathcal{L}_{ml}$, and $\mathcal{L}_{sp}$, respectively. 
We initially train the QG model with multi-task learning (\textit{MTL-QG}) setup. We choose the best \textit{MTL-QG} model which achieves the highest BLEU score on the development dataset. The multihop-enhanced reward based QG model is trained by initializing the best \textit{MTL-QG} model parameters.

\begin{table*}[h] \small
\centering
\begin{tabularx}{\textwidth}{X}
\hline
\textbf{Document (1):}  \textbf{(a)} \textcolor{blue}{ after } \textcolor{red}{bedřich smetana}\textcolor{blue} { , he was the second czech composer to achieve worldwide recognition .} ...
\\
\textbf{Document (2):} \textbf{(a)}
\textcolor{blue}{ concert at the end of summer ( czech : koncert na konci léta ) is 1980 czechoslovak historical film . } ... \\
\textbf{Target Answer:}  \textcolor{red}{bedřich smetana}\\
\textbf{Reference:}\textit{ which czech composer achieved worldwide recognition before the subject of " concert at the end of summer " ?}
\textbf{with only Rouge-L reward:} who was the composer of the composer of concert at the end of summer ? 
\\
\textbf{with Rouge-L and MER:} 
what was the \textcolor{forestgreen(traditional)}{\textbf{second czech composer}} to achieve worldwide recognition for the composer of the concert at the end of summer ? \\
\hline

\hline
\textbf{Document (1):} 
\textbf{(a)}  \textcolor{blue}{seedley railway station is a disused station located in the seedley area of pendleton , salford , on the liverpool and manchester railway . } ...
\\
\textbf{Document (2):}
\textbf{(a)} \textcolor{blue}{ pendleton is an inner city area of salford in greater manchester , england .} ...\\
\textbf{Target Answer:} \textcolor{red}{england}\\
\textbf{Reference:} \textit{seedley railway station is a disused station located in the seedley area of pendleton , is an inner city area of salford in greater manchester , in which country ?}\\
\textbf{with only Rouge-L reward:} : seedley railway station is located in a city area of salford in what country ?\\
\textbf{with Rouge-L and MER:} seedley railway station is a \textcolor{forestgreen(traditional)}{\textbf{disused station located}} in a city area of salford in greater manchester , in which country ?\\

\hline
\end{tabularx}
\caption{\small{Sample questions, where our proposed reward MER based model  generating better questions than only Rouge-L reward. The additional included information in the generated questions are shown in \textcolor{forestgreen(traditional)}{Green}}}
\label{tab:qg_example_reward}
\vspace{-1.0 em}
\end{table*}

\subsection{Dataset} \label{dataset}
We use the \texttt{HotPotQA} \cite{hotpot} dataset to evaluate our methods. This dataset consists of over $113$k Wikipedia-based question-answer pairs, with each question requiring multi-step reasoning across multiple supporting documents to infer the answer. While there exists other multi-hop datasets \cite{qangaroo,complex-web-qa-dataset}, only \texttt{HotPotQA} dataset provides the sentence-level labels to locate the supporting facts in the list of documents. The ground-truth information of the supporting facts facilitates stronger supervision for tracing the multi-step reasoning chains across the documents used to link the question with the answer. Our approach utilizes the ground-truth supporting facts information (only at the time of training) to form a better input representation and reinforcing desired behavior through multi-task learning and adaptive self-critical RL framework, respectively.

Each question in \texttt{HotPotQA} is associated with $10$ documents and the span information of the answer and supporting facts in these documents. However, only two of these documents effectively contain all the supporting facts and the ground-truth answer. The average number of supporting facts associated with a question is $2.38$ (these are all present in the $2$, out of $10$ documents that contain answer and all the supporting facts). The average length of a question and a document in \texttt{HotPotQA} are $21.82$ and $198.3$, respectively. In the pre-processing stage, we remove all the documents that do not contain an answer or supporting facts. Further, we remove all the comparison based question-answer pairs that have a `\textit{yes}' or `\textit{no}' answer. We then combine the training set ($90,564$) and development set ($7,405$) and randomly split the resulting data, with 80\% for training, 10\% for development, 10\% for testing. We call this split as training, development and test dataset of \texttt{HotPotQA}. We ensure that the proportion of difficulty level (easy, medium, hard) of questions was nearly uniform in the train, dev, and test sets.
\subsection{Hyperparameters}
In our experiments, we use the same vocabulary for both the encoder and decoder. Our vocabulary consists of the top 50,000 frequent words from the training data. We use the development dataset for hyper-parameter tuning. Pre-trained GloVe embeddings \cite{pennington2014glove} of dimension $300$ are used in the document encoding step. The hidden dimension of all the LSTM cells is set to $512$. Answer tagging features and supporting facts position features are embedded to 3-dimensional vectors. The dropout \cite{srivastava2014dropout} probability $p$ is set to $0.3$. The beam size is set to $4$ for beam search. We initialize the model parameters randomly using a Gaussian distribution with Xavier scheme \cite{Glorot10understandingthe}.

We first pre-train the network by minimizing only the maximum likelihood (ML) loss discussed in Eq \ref{eq:ml-loss}. Next, we initialize our model with the pre-trained ML weights and train the network with the mixed-objective learning function  as in Eq \ref{eq:mixed-loss}. The following values of hyperparameters were used to generate the results : (i) $\gamma_1=0.99$, $\gamma_2=0.01$, $\gamma_3=0.1$,  (ii) $d_1=300$,  $d_2=d_3=3$, (iii) $\alpha=0.9, \beta= 10$, $h=5000$.  Adam \cite{kingma2014adam} optimizer is used to train the model with (i) $ \beta_{1} = 0.9 $, (ii) $ \beta_{2} = 0.999 $, and (iii) $ \epsilon=10^{-8} $. For MTL-QG training, the initial learning rate is set to $0.01$. For our proposed model training the learning rate is set to $0.00001$. We also apply gradient clipping \cite{Pascanu:2013:DTR:3042817.3043083} with range $ [-5, 5] $. Based on grid-search, mini-batch size of 16 results in chosen for quick training and fast convergence. To train the supporting facts prediction model, we use the same dataset split as discussed in the Section \ref{dataset} and follow the hyper-parameter setting as discussed in \newcite{hotpot}.  The optimal beam size =$4$ is obtained from the model performance on the development dataset. 

\subsection{Evaluation}\label{eval}
We conduct experiments to evaluate the performance of our proposed and other QG methods using the evaluation metrics: BLEU-1, BLEU-2, BLEU-3, BLEU-4 \cite{papineni2002bleu}, ROUGE-L \cite{lin2004rouge} and METEOR \cite{Lavie:2009:MMA:1743627.1743643}. 
\paragraph{Metric for MultiHoping in QG:}
To assess the multi-hop capability of the question generation model, we introduce additional metric \emph{SF coverage}, which measures in terms of F1 score. This metric is similar to MultiHop-Enhanced Reward, where we use the question-aware supporting facts predictions network that takes the generated question and document list as input and predict the supporting facts. 
F1 score measures the average overlap between the predicted and ground-truth supporting facts as computed in \cite{hotpot}.
\section{Results and Analysis}
We first describe some variants of our proposed MultiHop-QG model.
\begin{enumerate}
\item \textbf{SharedEncoder-QG}: This is an extension of the \texttt{NQG} model \cite{zhou2017neural-qg} with shared encoder for QG and answer-aware supporting fact predictions tasks. This model is a variant of our proposed model, where we encode the document list using a two-layer Bi-LSTM which is shared between both the tasks. The input to the shared Bi-LSTM is word and answer encoding as shown in Eq. \ref{lstm-1st-layer}. The decoder is a single-layer LSTM which generates the multi-hop question. 
\item \textbf{MTL-QG}: This variant is similar to the \texttt{SharedEncoder-QG}, here we introduce another Bi-LSTM layer which takes the question, answer and supporting fact embedding as shown in Eq. \ref{lstm-2nd-layer}.
\end{enumerate}
\begin{figure}
\begin{minipage}{\textwidth}
  \begin{minipage}[b]{0.55\textwidth}
\resizebox{\linewidth}{!}{
\begin{tabular}{c|l|l|l}
\hline
\textbf{Model}  & \textbf{BLEU-4} & \textbf{ROUGE-L}  & \textbf{SF Coverage}  \\ \hline
\textit{NQG} \cite{zhou2017neural-qg} & 19.55 &  36.63 & 61.55 \\ 
\textit{SharedEncoder-QG} (NQG + Shared Encoder) & 20.64  &37.18 & 65.46\\
\textit{MTL-QG (SharedEncoder-QG + SF)} & 21.28  &37.55 & 70.11\\ 
\textit{MTL-QG + Rouge-L} & 22.83 &  39.41 & 71.27\\ \hline
\begin{tabular}[c]{@{}c@{}}Proposed Model\\ (MTL-QG + SF + Rouge-L + MER) \end{tabular}  & \textbf{23.57 }&  \textbf{39.68} & \textbf{74.37}\\  \hline \hline
\end{tabular}%
}
\captionof{table}{ A relative performance (on test dataset of \texttt{HotPotQA}
) of different variants of the proposed method, by adding one model component. }
\label{ablation}
\end{minipage}
\hfill
 \begin{minipage}[b]{0.4\textwidth}
    \centering
   \resizebox{\linewidth}{!}{%
    \begin{tabular}{lcccc} 
    \toprule
     Model & Naturalness & Difficulty & SF Coverage \\\midrule
    
\textit{NQG}  & 3.20 & 2.42 & 73.12  \\
Proposed  & \textbf{3.47} & \textbf{3.21} & \textbf{83.04}  \\
    \bottomrule
    \end{tabular}
    }
 \captionof{table}{Human evaluation results for our proposed approach and the NQG model. Naturalness and difficulty are rated on a 1--5 scale and SF coverage is in percentage (\%).}
\label{tab:human}
\end{minipage}
\end{minipage}
\end{figure}
\indent The automatic evaluation scores of our proposed method, baselines, and state-of-the-art single-hop question generation model on the \texttt{HotPotQA} test set are shown in Table \ref{main-results}. We also show 
the additional results on the development dataset in the \textbf{Appendix}. The performance improvements with our proposed model over the baselines and state-of-the-arts are statistically significant\footnote{We follow the bootstrap test \cite{efron1994introduction} using the setup provided by \newcite{P18-1128}.} as $(p <0.005)$. For the question-aware supporting fact prediction model (c.f. \ref{mer-reward}), we obtain the F1 and EM scores of $84.49$ and $44.20$, respectively, on the \texttt{HotPotQA} development dataset. We can not directly compare the result ($21.17$ BLEU-4) on the HotPotQA dataset reported in \newcite{nema-etal-2019-lets} as their dataset split is different and they only use the ground-truth supporting facts to generate the questions. 

We also measure the multi-hopping in terms of SF coverage and reported the results in Table \ref{main-results} and Table \ref{ablation}. We achieve skyline performance of $80.41$ F1 value on the ground-truth questions of the test dataset of HotPotQA. 
\subsection{Quantitative Analysis}
Our results in Table \ref{main-results} are in agreement with \cite{ans-focused-qg,para-level-max-out-qg,zhou2017neural-qg}, which establish the fact that providing the answer tagging features as input leads to considerable improvement in the QG system's performance. Our \textit{SharedEncoder-QG} model, which is a variant of our proposed MultiHop-QG model outperforms all the baselines state-of-the-art models except \textit{Semantic-Reinforced}. 
The proposed \textit{MultiHop-QG} model achieves the absolute improvement of $4.02$ and $3.18$ points compared to \textit{NQG} and \textit{Max-out Pointer} model, respectively, in terms of BLEU-4 metric.  \\
\indent To analyze the contribution of each component of the proposed model, we perform an ablation study reported in Table \ref{ablation}. Our results suggest that providing multitask learning with shared encoder helps the model to improve the QG performance from $19.55$ to $20.64$ BLEU-4. Introducing the supporting facts information obtained from the answer-aware supporting fact prediction task further improves the QG performance from $20.64$ to $21.28$ BLEU-4. Joint training of QG with the supporting facts prediction provides stronger supervision for identifying and utilizing the supporting facts information. In other words, by sharing the document encoder between both the tasks, the network encodes better representation (supporting facts aware) of the input document. Such presentation is capable of efficiently filtering out the irrelevant information when processing multiple documents and performing multi-hop reasoning for question generation. Further, the MultiHop-Enhanced Reward (MER) with Rouge reward provides a considerable advancement on automatic evaluation metrics. 
We also perform the experiment with different beam size and reported the performance of QG model is given in the \textbf{Appendix}. 
\subsection{Qualitative Analysis}
We have shown the examples in Table \ref{tab:qg_example_reward}, where our proposed reward 
assists the model to maximize the uses of all the supporting facts to generate better human alike questions. In the first example, Rouge-L reward based model ignores the information `\textit{second czech composer}' from the first supporting fact, whereas our MER reward based proposed model considers that to generate the question. Similarly, in the second example, our model considers the information `\textit{disused station located}' from the supporting fact where the former model ignores it while generating the question. We also compare the questions generated from the \textit{NQG} and our proposed method with the ground-truth questions. These questions with additional generated questions are given in the \textbf{Appendix}.
\paragraph{Human Evaluation: }
For human evaluation, we directly compare the performance of the proposed approach with NQG model. We randomly sample 100 document-question-answer triplets from the test set and ask four professional English speakers to evaluate them. We consider three modalities: \emph{naturalness}, which indicates the grammar and fluency; \emph{difficulty}, which measures the document-question syntactic divergence and the reasoning needed to answer the question, and \emph{SF coverage} similar to the metric discussed in Section \ref{exp-setup} except we replace the supporting facts prediction network with a human evaluator and we measure the relative supporting facts coverage compared to the ground-truth supporting facts.
measure the relative coverage of supporting facts in the questions 
with respect to the ground-truth supporting facts. \emph{SF coverage} provides a measure of the extent of supporting facts used for question generation. 
For the first two modalities, evaluators are asked to rate the performance of the question generator on a 1--5 scale (5 for the best). To estimate the \emph{SF coverage} metric, the evaluators are asked to highlight the supporting facts from the documents based on the generated question. 
We reported the average scores of all the human evaluator for each criteria in Table \ref{tab:human}.
The proposed approach is able to generate better questions in terms of \textit{Difficulty}, \textit{Naturalness} and \textit{SF Coverage} when compared to the \textit{NQG} model. 
\section{Conclusion}
In this paper, we have introduced the multi-hop question generation task, which extends the natural language question generation paradigm to multiple document QA. Thereafter, we present a novel reward formulation to improve the multi-hop question generation using reinforcement and multi-task learning frameworks. Our proposed method performs considerably better than the state-of-the-art question generation systems on \texttt{HotPotQA} dataset. We also introduce SF Coverage, an evaluation metric to compare the performance of question generation systems based on their capacity to accumulate information from various documents. Overall, we propose a new direction for question generation research with several practical applications. In the future, we will be focusing on to improve the performance of multi-hop question generation without any strong supporting facts supervision.

\section*{Acknowledgement}
Asif Ekbal gratefully acknowledges the Young Faculty Research Fellowship (YFRF) Award supported by the Visvesvaraya PhD scheme for Electronics and IT, Ministry of Electronics and Information Technology (MeitY), Government of India, and implemented by Digital India Corporation (formerly Media Lab Asia).
\bibliographystyle{coling}
\bibliography{coling2020}

\appendix
\begin{table*}[h] 
\begin{center}
\resizebox{0.95\textwidth}{!}{%
\begin{tabular}{l|l|l|l|l|l|l|l}
\hline
\textbf{Model} & \textbf{BLEU-1} & \textbf{BLEU-2} & \textbf{BLEU-3} & \textbf{BLEU-4} & \textbf{METEOR} & \textbf{ROUGE-L} & \textbf{SF Coverage} \\ \hline
\textit{s2s} \cite{learning-to-ask} & 35.00 & 22.57 & 16.80 & 13.29 & 14.90 &  29.12&60.31      \\ 
\textit{s2s+copy} & 33.52 & 23.47 & 18.46 & 15.27 & 18.81 & 30.66 &  60.45\\ 
\textit{s2s+answer} & 39.72 & 27.25 & 20.82 & 16.62 & 17.58 & 33.94 &   61.14\\
\textit{NQG} \cite{zhou2017neural-qg} & 42.67 & 30.53 & 23.86 & 19.42 & 20.69 & 36.79 & 61.68\\ 
\textit{ASs2s} \cite{kim2018improving} & 42.18 & 29.97 & 23.30 & 18.87 & 21.27 & 37.34 & 63.98\\ 
\textit{Max-out Pointer} \cite{para-level-max-out-qg} & 42.05 & 30.48 & 24.29 & 20.17 & 20.17 & 34.93 &64.19 \\ 

\textit{Semantic-Reinforced} \cite{zhang-bansal-2019-addressing} & 43.84 & 32.84. & 26.02 & 21.43 & 23.47 & 39.17 & 68.25\\
\hline
\textit{SharedEncoder-QG} (Ours)& 44.51 & 31.95 & 25.04 & 20.40 & 21.66 &37.83 & 65.22  \\
\textit{MTL-QG} (Ours) & 44.10 & 32.39 & 25.94 & 21.58 & 21.76 &37.77 & 69.24 \\ \hline
Proposed Model & \textbf{46.95} & \textbf{38.85} &\textbf{ 27.83} & \textbf{23.25 }& \textbf{22.93} & \textbf{39.74} & \textbf{73.82} \\ \hline
\hline

\end{tabular}%
}
\end{center}
\caption{Automatic evaluation scores of the proposed approach and the baselines on the development set of \texttt{HotPotQA}.}
\label{main-results-dev}
\end{table*}

\begin{table}[h] 
\centering
\resizebox{0.7\columnwidth}{!}{%
    \begin{tabular}{lcccc} 
    \toprule
     Model & BLEU-4 & ROUGE-L & METEOR & SF Coverage \\\midrule
    
{Proposed Model}  & $\textbf{23.57}$ & $\textbf{39.68}$ & $22.88$ &\textbf{73.82} \\ \hline
\quad {w/ Beam Size 3} & $23.29$ & $39.67$ & $22.69$ & 73.47\\
\quad  {w/ Beam Size 5} & 23.47 & 39.60 & \textbf{22.98} & \textbf{73.82}\\
\quad  {w/ Beam Size 7}  & 23.37 & 39.54 & 23.13 &73.54 \\
\quad  {w/ Beam Size 10}  & 23.12 & 39.42 & 23.18 &72.94 \\
    \bottomrule
    \end{tabular}}
  \caption{Performance of question generation on the HotPotQA test dataset by varying beam size.}
\label{tab:beam-res}
\end{table}

\begin{table*}[!h] \small
\centering
\begin{tabularx}{\textwidth}{X} 
\hline

\textbf{Document (1):} 
\textbf{(a)}  \textcolor{blue}{the}  \textcolor{red}{m6 motorway} \textcolor{blue}{runs from junction 19 of the m1 at the catthorpe interchange , near rugby via birmingham then heads north , passing stoke - on - trent , liverpool , manchester , preston , lancaster , carlisle and terminating at the gretna junction ( j45 ) .}
\textbf{(b)} here , just short of the scottish border it becomes the a74(m ) which continues to glasgow as the m74 .\\
\textbf{Document (2):}
\textbf{(a)} \textcolor{blue}{ shap is a linear village and civil parish located among fells and isolated dales in eden district , cumbria , england .}
\textbf{(b)} the village lies along the a6 road and the west coast main line , and is near to the m6 motorway .
\textbf{(c)}  it is situated 10 mi from penrith and about 15 mi from kendal , in the historic county of westmorland .

\textbf{Target Answer:} \textcolor{red}{m6 motorway}\\
\textbf{Reference:} \textit{what motoway runs from junction 19 of the m1 and is near the linear village shap ?}\\
\textbf{NQG:} what motorway runs from junction 19 of the m1 at the catthorpe interchange ?  \\
\textbf{Proposed:} what motorway runs from junction 19 of the m1 at the catthorpe interchange and is near shap ?\\

\hline
\textbf{Document (1):}  \textbf{(a)} \textcolor{blue}{the east mamprusi district is one of the twenty ( 20 ) districts in the northern region of north ghana .} \textbf{(b)} \textcolor{blue}{the capital is} \textcolor{red}{gambaga} .\\
\textbf{Document (2):} \textbf{(a)}
 \textcolor{blue}{shienga ( shinga ) is a village in east mamprusi district , of the northern region of ghana .}
 \textbf{(b)} it lies at an elevation of 349 meters near the right ( southern ) bank of the white volta .\\
\textbf{Target Answer:}  \textcolor{red}{gambaga}\\
\textbf{Reference:} \textit{what is the capitol of the district that also includes the village of shienga ?} \\
\textbf{NQG:} what is the capital of the east mamprusi district ?
  \\
\textbf{Proposed:} what is the capital of the district in which shienga is located ? \\
\hline
\textbf{Document (1):} 
\textbf{(a)} \textcolor{blue}{robert clinton smith ( born march 30 , 1941 ) is an american politician who served as a member of the united states house of representatives for new hampshire 's 1st congressional district from 1985 to 1990 and the state of new hampshire in the united states senate from 1990 to 2003 .} \\
\textbf{Document (2):}
\textbf{(a)} new hampshire 's 1st congressional district covers the southeastern part of new hampshire .
\textbf{(b)} \textcolor{blue}{the district consists of \textcolor{red}{three} general areas : greater manchester , the seacoast and the lakes region .}  \\
\textbf{Target Answer:} \textcolor{red}{three}\\
\textbf{Reference:}\textit{ bob smith served as a member of the united states house of representatives for a district that consists of how many general areas ?}\\
\textbf{NQG:} bob smith served as a member of how many general areas ? \\
\textbf{Proposed:} bob smith served as a member of the united states house of representatives for a district that consists of how many general areas ?\\
\hline
\textbf{Document (1):} 
\textbf{(a)}  \textcolor{blue}{leanne rowe ( born 1982 ) is an english actress and singer , known for portraying nancy in " oliver twist " , may moss in " lilies " and baby in " dirty dancing : the classic story on stage " . }\\
\textbf{Document (2):}
\textbf{(a)} \textcolor{blue}{ oliver twist is a 2005 drama film directed by} \textcolor{red}{roman polanski}
\textbf{(b)} the screenplay by ronald harwood is based on the 1838 novel of the same name by charles dickens .

\textbf{Target Answer:} \textcolor{red}{roman polanski}\\
\textbf{Reference:} \textit{who directed the 2005 film in which leanne rowe portrayed nancy ?}\\
\textbf{NQG:} who directed the film in which leanne rowe played nancy ?  \\
\textbf{Proposed:} who directed the 2005 drama film in which leanne rowe played nancy ?\\
\hline

\textbf{Document (1):} \textbf{(a)}  \textcolor{blue}{not without laughter is the debut novel by langston hughes published in 1930 .} \\
\textbf{Document (2):} \textbf{(a)}
 \textcolor{blue}{james mercer langston hughes ( february 1 , 1902 – may 22 , 1967 ) was an \textcolor{red}{american} poet .}
 \textbf{(b)} he was a social activist , novelist , playwright , and columnist from joplin , missouri .\\
\textbf{Target Answer:}  \textcolor{red}{american}\\
\textbf{Reference:} \textit{what was the nationality of the author of " not without laughter " ?} \\
\textbf{NQG:} not without laughter is a novel by a man of what nationality ?  \\
\textbf{Proposed:} not without laughter is the debut novel by the poet of what nationality ?\\
\hline
\hline

\end{tabularx}
\caption{\small{Samples generated by multi-hop question generation approach. In each document, the supporting facts are shown in \textcolor{blue}{blue} and the target answer is in \textcolor{red}{red}. }}

\label{tab:qg_example_basic}
\end{table*}
\begin{table*}[h]
\begin{center}
\resizebox{0.95\textwidth}{!}{

\begin{tabular}{c|l|l|l|l|l|l|l}
\hline
\textbf{Model} & \textbf{BLEU-1} & \textbf{BLEU-2} & \textbf{BLEU-3} & \textbf{BLEU-4} & \textbf{METEOR} & \textbf{ROUGE-L} & \textbf{SF Coverage}  \\ \hline
\textit{NQG} \cite{zhou2017neural-qg} & 42.67 & 30.53 & 23.86 & 19.42 & 20.69 & 36.79 &61.68 \\
\textit{SharedEncoder-QG} (NQG + Shared Encoder) & 44.51 & 31.95 & 25.04 & 20.40 & 21.66 &37.83 & 65.22\\
\textit{MTL-QG (SharedEncoder-QG + SF)} & 44.10 & 32.39 & 25.94 & 21.58 & 21.76 &37.77 & 69.24\\
\textit{MTL-QG + BLEU} & 47.29 & 33.12 & 26.14 & 22.46 & 22.19 & {39.93} &70.15  \\
\textit{MTL-QG + Rouge-L} & 47.16 & 34.81 & 27.72 & 22.83 & 22.78 & {40.04} & 70.87  \\ \hline
\begin{tabular}[c]{@{}c@{}}Proposed Model\\ (MTL-QG + SF + Rouge-L + MER) \end{tabular} & {\textbf{46.95}} & {\textbf{38.85}} & {\textbf{27.83}} & {\textbf{23.25} }& {\textbf{22.93}} & \textbf{39.74} &\textbf{73.82} \\  \hline \hline
\end{tabular}%
}
\end{center}
\caption{ A relative performance (on development dataset of \texttt{HotPotQA}
) of different variants of the proposed method, by adding one model component. }

\label{ablation-dev}
\end{table*}

\begin{table*}[!h] \small
\centering
\begin{tabularx}{\textwidth}{X} 
\hline
\textbf{Document (1):} 
\textbf{(a)} \textcolor{blue}{orange is a town in grafton county , new hampshire , united states .  the population was 331 at the 2010 census .} \\
\textbf{Document (2):}
\textbf{(a)} \textcolor{blue}{cardigan mountain state park is a \textcolor{red}{5655 acre} state park in orange , new hampshire .}
\textbf{(b)} the park is free to use , open year - round , and offers a hiking trail up to the 3,121-foot treeless granite summit of mount cardigan . \textbf{(c)}
there are picnic facilities .
\\
\textbf{Target Answer:} \textcolor{red}{5655 acre}\\
\textbf{Reference:} \textit{how big is the state park located in grafton county , new hampshire ?}\\
\textbf{NQG:} how acres is the state park located in orange , new hampshire ? \\
\textbf{with only Rouge-L reward:} how many acres is the park in which orange , new hampshire is located ?	 \\
\textbf{with Rouge-L reward and MER:} how many acres is the state park in which town in grafton county , new hampshire ?\\
\hline
\textbf{Document (1):} 
\textbf{(a)}  \textcolor{blue}{buddy stephens is an american football coach who is currently the head coach at east mississippi community college , where he has won three njcaa national championships and coached players such as chad kelly and} \textcolor{red}{john franklin iii} .
\textbf{(b)} with an overall record of 87–12 , stephens has a higher winning percentage ( .879 ) than the njcaa all - time leader ( butler cc 's troy morrell at 154–22 for .875 ) , but has not yet coached the required 100 games to appear on the list .\\
\textbf{Document (2):}
\textbf{(a)}  john franklin iii is an american football wide receiver for the florida atlantic owls football . \\
\textbf{(b)} \textcolor{blue}{he formerly played for florida state university , east mississippi community college and auburn university .}\\
\textbf{Target Answer:} \textcolor{red}{john franklin iii}\\
\textbf{Reference:} \textit{what player did buddy stephens coach who went on to play for florida state university , east mississippi community college and auburn university ?}\\
\textbf{NQG:} buddy stephens is an american football coach who is currently the head coach at east mississippi community college , where he has won three njcaa national championships and coached players such as chad kelly and which american football wide receiver for the florida atlantic owls football ?  \\
\textbf{with only Rouge-L reward: } what american football wide receiver formerly played for the florida atlantic owls football wide receiver for the florida atlantic owls football , coached buddy stephens ?\\ 
\textbf{with Rouge-L reward and MER:} what is the name of the chad kelly  american football coach who formerly formerly played for florida state university , east mississippi community college and auburn university ?\\

\hline
\textbf{Document (1):} 
\textbf{(a)}  \textcolor{blue}{seedley railway station is a disused station located in the seedley area of pendleton , salford , on the liverpool and manchester railway . } 
\textbf{(b)} 
it was opened on 1 may 1882 and closed on 2 january 1956 . 
\textbf{(c)} parts of the station wall can still be seen but part of the trackbed has been covered over following the construction of the m602 motorway .
\\
\textbf{Document (2):}
\textbf{(a)} \textcolor{blue}{ pendleton is an inner city area of salford in greater manchester , england .} \\
\textbf{(b)} it is about 2 mi from manchester city centre .  
\textbf{(c)} the a6 dual carriageway skirts the east of the district .\\
\textbf{Target Answer:} \textcolor{red}{england}\\
\textbf{Reference:} \textit{seedley railway station is a disused station located in the seedley area of pendleton , is an inner city area of salford in greater manchester , in which country ?}\\
\textbf{NQG:} what country does [UNK] railway station and pendleton , greater manchester have in common ? \\
\textbf{with only Rouge-L reward} : seedley railway station is located in a city area of salford in what country ?\\
\textbf{with Rouge-L reward and MER:} seedley railway station is a disused station located in a city area of salford in greater manchester , in which country ?\\

\hline
\end{tabularx}
\caption{\small{Sample questions, where our proposed MER based reward model  generating better questions than only Rouge-L reward and NQG model.}}

\label{tab:qg_example-1}
\end{table*}



\section{Question-Aware Supporting Fact Prediction Network}
The network uses a convolutional network to obtain a character-based word embedding, which is concatenated with a pre-trained word embedding, and followed by a recurrent layer to encode the contextual information. The same is applied to both question and document. Thereafter, a bi-directional attention layer \cite{Seo2017Bidirectional} is employed to fuse the representations of the question and the documents. After using another recurrent layer, we add a self-attention layer \cite{wang2017gated} followed by a residual connection. 
For each candidate sentence in the document list, we concatenate the output of the self-attention layer at the first and last positions, and use a binary linear classifier to predict the probability that the current sentence is a supporting fact. 
\section{Analysis}
We compare the questions generated from the \textit{NQG} and our proposed method with the ground-truth questions, examples of which are shown in Table \ref{tab:qg_example_basic}. 
The very first example 
focuses on both the supporting facts, while the NQG model only uses one supporting fact. In the second example, the NQG system generates the question by considering the target answer and a single document. Unlike, our proposed model uses the relation information (\textit{shinga} is a village in \textit{east mamprusi district}) that exists between the entities across the document. It is because our proposed model focuses on all the supporting facts for question generation. We have also shown the examples in Table \ref{tab:qg_example-1}, where our proposed reward 
assists the model to maximize the uses of all the supporting facts to generate better human alike questions.

\end{document}